\documentclass[times,twocolumn,final]{elsarticle}

\usepackage{ipcai_arxiv}
\usepackage{graphicx}%
\usepackage{multirow}%
\usepackage{amsmath,amssymb,amsfonts}%
\usepackage{amsthm}%
\usepackage{mathrsfs}%
\usepackage[title]{appendix}%
\usepackage{xcolor}%
\usepackage{textcomp}%
\usepackage{manyfoot}%
\usepackage{booktabs}%
\usepackage{algorithm}%
\usepackage{algorithmicx}%
\usepackage{algpseudocode}%
\usepackage{listings}%
\usepackage{multirow}
\usepackage{pifont}%
\usepackage{adjustbox}
\usepackage{graphicx}
\usepackage{hyperref}
\usepackage{float}
\usepackage{placeins}
\usepackage{fancyhdr}
\usepackage{dsfont}
\usepackage{threeparttable}

\begin{document}
\fancypagestyle{firstpagestyle}{
    \fancyhf{} 
    \renewcommand{\headrulewidth}{0pt} 
    \fancyhead{} 
    \fancyfoot{} 
    \fancyhead[CO]{\em \fontsize{9pt}{8pt}\selectfont This article has been accepted to the International Conference on Information Processing in Computer-Assisted Interventions, 2025.}
}

\fancypagestyle{default}{
    \fancyhf{}
    \fancyhead[R]{\thepage} 
    \renewcommand{\headrulewidth}{0.4pt} 
    \fancyhead[LO]{\em \fontsize{9pt}{8pt}\selectfont This article has been accepted to the International Conference on Information Processing in Computer-Assisted Interventions, 2025.}
}



\title{Early Operative Difficulty Assessment in Laparoscopic Cholecystectomy via Snapshot-Centric Video Analysis}
\author[1,2]{Saurav \snm{Sharma} \fnref{corresp}}
\fntext[corresp]{Corresponding author: \texttt{ssharma@unistra.fr}}
\author[2,4]{Maria \snm{Vannucci}}
\author[1,2]{Leonardo Pestana \snm{Legori}}
\author[3]{Mario \snm{Scaglia}}
\author[5]{Giovanni Guglielmo \snm{Laracca}}
\author[2,8]{Didier \snm{Mutter}}
\author[6,7]{Sergio \snm{Alfieri}}
\author[2,6,7]{Pietro \snm{Mascagni} \fnref{coauth}}
\author[1,2]{Nicolas \snm{Padoy} \fnref{coauth}}
\fntext[coauth]{shared last authorship}

\address[1]{University of Strasbourg, CNRS, INSERM, ICube, UMR7357, France}
\address[2]{IHU Strasbourg, Strasbourg, France}
\address[3]{Università degli Studi di Milano}
\address[4]{General Surgery Department, University of Torino, Turin, Italy}
\address[5]{Department of Medical Surgical Science and Translational Medicine, Sant'Andrea Hospital, Sapienza University of Rome, Rome, Italy}
\address[6]{Fondazione Policlinico Universitario A. Gemelli IRCCS, Rome, Italy}
\address[7]{Università Cattolica del Sacro Cuore, Rome, Italy}
\address[8]{University Hospital of Strasbourg, France}

\received{XXX}
\finalform{XXX}
\accepted{XXX}
\availableonline{XXX}
\communicated{XXX}

\begin{abstract}
\textbf{Purpose}:
Laparoscopic cholecystectomy (LC) operative difficulty (LCOD) is highly variable and influences outcomes. Despite extensive LC studies in surgical workflow analysis, limited efforts explore LCOD using intraoperative video data. Early recognition of LCOD could allow prompt review by expert surgeons, enhance operating room (OR) planning, and improve surgical outcomes. 

\noindent\textbf{Methods}:
We propose the clinical task of early LCOD assessment using limited video observations. We design SurgPrOD, a deep learning model to assess LCOD by analyzing features from global and local temporal resolutions (snapshots) of the observed LC video. Also, we propose a novel snapshot-centric attention (SCA) module, acting across snapshots, to enhance LCOD prediction. We introduce the CholeScore dataset, featuring video-level LCOD labels to validate our method.

\noindent\textbf{Results}:
We evaluate SurgPrOD on 3 LCOD assessment scales in the CholeScore dataset. On our new metric assessing early and stable correct predictions, SurgPrOD surpasses baselines by at least 0.22 points. SurgPrOD improves over baselines by at least 9 and 5 percentage points in F1 score and top1-accuracy, respectively, demonstrating its effectiveness in correct predictions.

\noindent\textbf{Conclusion}:
We propose a new task for early LCOD assessment and a novel model, SurgPrOD analyzing surgical video from global and local perspectives. Our results on the CholeScore dataset establishes a new benchmark to study LCOD using intraoperative video data.
\\

\end{abstract}

\maketitle
\thispagestyle{firstpagestyle}

\section{Introduction}\label{intro}

Laparoscopic cholecystectomy (LC), the gold standard procedure for the gallbladder excision, is central to surgical workflow analysis in developing context-aware decision support systems~\cite{maier2022surgical,vercauteren2019cai4cai}. These systems are designed to assist the surgeons and potentially improve patient outcomes through data driven approaches. Recent advancements have focused on learning robust surgical scene representations from intraoperative video data in the operating room (OR) to analyze key elements of surgical procedures. Formulated as deep learning tasks, these elements include recognition of phase~\cite{twinanda2016endonet}, steps~\cite{Lavanchy2024}, tool-tissue interactions~\cite{nwoye2021rendezvous,sharma2022rendezvous,sharma2023surgical}, anatomical structures~\cite{murali2023latent}, and crucial process measures such as the Critical View of Safety (CVS)~\cite{mascagni2022artificial}.

Still, these works do not consider laparoscopic cholecystectomy operative difficulty (LCOD). LCOD is known to be highly variable and affects surgical and patients outcomes. Studying LCOD is complex due to the intrinsic interplay among patients factors, disease severity, and surgical performance. However, since LC is performed by most general surgeons, often in their learning curve, and across hospitals, assessing and predicting LCOD could significantly enhance patient stratification, optimize allocation of expertise and resources, and improve outcomes.

\begin{figure*}[h!]
\centering
\includegraphics[width=0.70\textwidth]{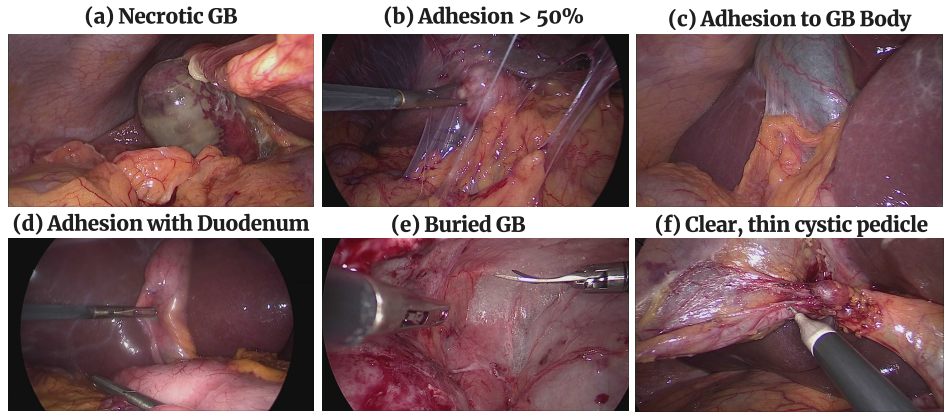}
\caption{CholeScore: Sample frames with the associated LCOD findings.} \label{example_lcod}
\end{figure*}

Our previous work~\cite{vannucci2022statistical} examined statistical models to preoperatively predict LCOD using variables such as demographics and ultrasound findings. While promising, these models predict a variety of operator dependent outcomes such as conversion to open surgery and operating time. Intraoperative findings such as adhesions, gallbladder inflammation, and gallstones provide more operator-independent visual cues to assess LCOD. Intraoperative assessment scales (IOAS) like the Parkland~\cite{madni2018parkland}, Nassar~\cite{nassar1995laparoscopic}, and Sugrue~\cite{sugrue2015grading} scales analyze a group of these intraoperative findings to categorize LCOD into distinct grades, offering a standardized framework for assessing surgical complexity. Figure~\ref{example_lcod} shows LC frames and their difficulty grades. Still, IOAS assessment via direct observation or video review requires expert surgeon's time, limiting IOAS use to the research setting. 

Motivated by these insights and the potential to inform clinicians about LCOD prior to the safety critical steps of the procedure, we introduce the novel and clinically relevant task of early LCOD assessment. This task provides a holistic LCOD assessment for the entire procedure by analyzing only the initial minutes of the intraoperative video, similar to early surgery prediction~\cite{kannan2019future}. We seek to address two key questions: $(1)$ \textit{How can we design a framework to automatically assess LCOD using limited video data?}, and $(2)$ \textit{what metrics to use for evaluating model performance on early assessment?}

To answer the first question, we propose \textbf{SurgPrOD} (\textbf{Surg}ical \textbf{Pr}edictor of \textbf{O}perative \textbf{D}ifficulty), a novel video-based deep learning method to assess LCOD using partial video data. Inspired by TemPr~\cite{stergiou2023wisdom}, SurgPrOD analyzes the observed portion of the surgical video at different temporal resolutions (\textit{snapshots}). It operates on one global and multiple non-overlapping local snapshots (Figure~\ref{fig:surgprod}), each with a fixed set of frames. SurgPrOD extracts visual features and generates class probabilities per snapshots. These are averaged across snapshots to produce a refined LCOD score. As the early assessment task is challenging, without temporal context, salient cues from local snapshots might be ignored due to averaging. We propose a snapshot-centric attention (\textbf{SCA}) module to facilitate semantic transfer between local snapshots of different temporal horizons, enhancing operative difficulty assessment.

To answer the second question, we observe that standard metrics like top1-accuracy and F1 score, while applicable to overall predictions, fail to capture a model's ability to accurately assess operative difficulty early and maintain stable predictions. This capability is crucial for decision support systems, as gains in overall predictions do not necessarily translate to performance in early assessments. To address this, we propose the \textbf{Earliness Stability (ES)} metric, which analyzes model predictions across all observation windows and computes a score indicating both earliness (how soon the correct class is predicted) and stability (consistency of the predictions over time). Additionally, we employ the Quadratic Weighted Cohen Kappa (QWK) metric, which penalizes larger deviations from the correct LCOD label. These metrics complement the traditional measures, offering nuanced evaluation of early assessment methods.

To benchmark our methods, we generate CholeScore, a unique dataset of $100$ videos annotated with video-level operative difficulty labels across the $3$ IOAS - Parkland~\cite{madni2018parkland}, Sugrue~\cite{sugrue2015grading}, and Nassar~\cite{nassar1995laparoscopic} scales. Based on the intraoperative findings, each scale categorizes the videos into grades of surgical difficulty. We evaluate SurgPrOD on the CholeScore dataset for the task of early LCOD assessment across $3$ IOAS, showcasing gains over baseline methods. Thanks to ES metric, we demonstrate our model's ability to accurately assess the LCOD early in the procedure.

We summarize our main contributions below:
\begin{itemize}
    \item We propose the novel and clinically meaningful task of LCOD assessment using limited video observations captured during the early stages of the procedure.
    \item We design a novel deep learning method SurgPrOD that predicts LCOD using global and local snapshots of the observed surgical video.
    \item We evaluate our model on 3 clinical LCOD assessment scales and report improvements over baseline methods.
    \item We propose a new metric measuring prediction earliness and stability.
\end{itemize}

\section{Methodology}
\subsection{Problem Setup}
Our goal is to assess the overall LCOD using partial observations from the start of surgical procedure. Given a video with $F$ frames, we define an observation window $w$ (in minutes), where $w \in [1, w_{max}]$ represents a portion of the surgical video. For each $w$ of increasing size, SurgPrOD analyzes the first $F_w$ ($F_w < F$) frames and for each IOAS outputs class probabilities in $\mathbb{R}^C$, where $C$ is the number of LCOD classes.

\begin{figure*}[h!]
\includegraphics[width=0.95\textwidth]{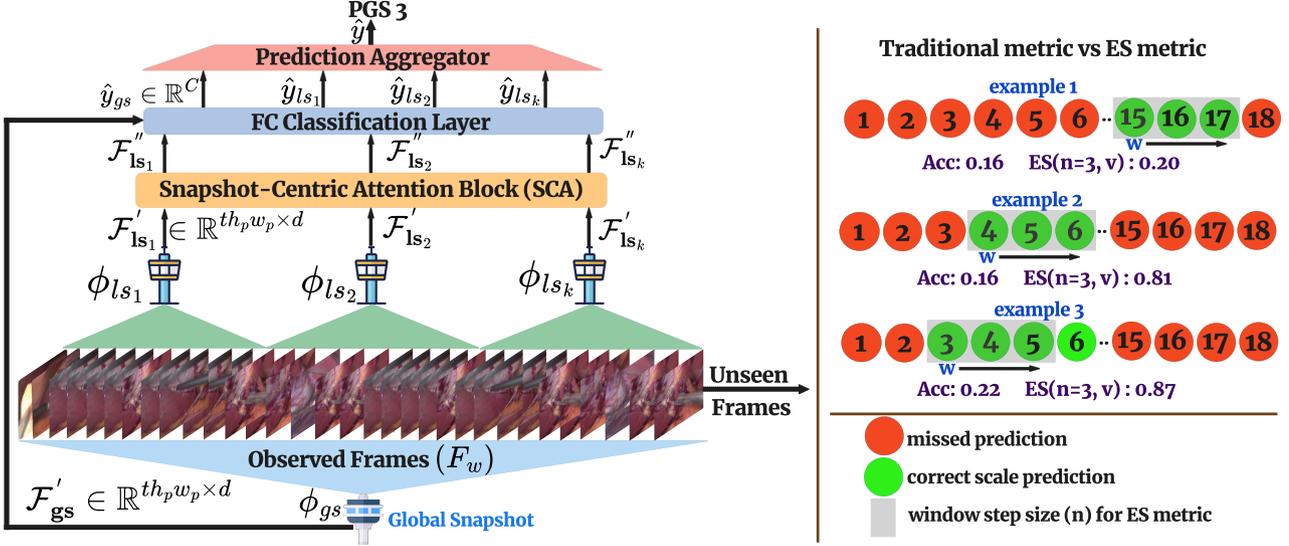}
\caption{\textbf{Model Overview}: (left) \textbf{SurgPrOD} inputs $F_{w}$ observed frames to generate a global snapshot $\mathbf{gs}$ and $k$ local snapshots $\mathbf{ls}_{k}$. MoCoV2~\cite{ramesh2023dissecting} features are extracted for each snapshot and processed through a transformer $\phi$. Snapshot Centric-Attention (SCA) enhances the $k$ local snapshot features to $\mathcal{F}^{''}_{\mathbf{{ls}_{k}}}$, and together with global snapshot features $\mathcal{F}^{'}_{\mathbf{gs}}$, inputs to a MLP layer to produce class logits and averaged to compute the LCOD class probabilities. (right) The Early Stability (ES) metric (Equation~\ref{es_metric}), addresses the limitations of traditional metrics by rewarding early (observation window $w$) and stable correct predictions (green circles) within a window step size (n=3, gray boxes). Circles represent observation windows.} \label{fig:surgprod}
\end{figure*}

\subsection{SurgPrOD}
Inspired by TemPr~\cite{stergiou2023wisdom}, we design SurgPrOD, a novel video-based architecture (Figure~\ref{fig:surgprod}). TemPr progressively samples frames across multiple scales of the observed video, extracts features to compute predictions that are aggregated to predict action in the video. In contrast to TemPr, SurgPrOD collects multiple sets of frames across different temporal resolutions, referred to as \textit{snapshots} for a window $w$ with $F_w$ frames. We create a global snapshot with $t$ randomly sampled frames from $F_w$, generating features that capture overall visual cues. For fine-grained scene features, we partition $F_w$ into $k$ fixed, non-overlapping local snapshots, each with $t$ randomly sampled frames. The $k$ local snapshot features are further refined using a snapshot-centric attention (SCA) module. Finally, we generate class logits from all snapshots, averaged to output a final prediction. We describe each component of SurgPrOD in detail.

\subsubsection{Backbone}
We utilize MoCov2~\cite{ramesh2023dissecting}, a self-supervised model trained on Cholec80~\cite{twinanda2016endonet} as a feature extractor, due to its proven robustness and generalization ability across multiple tasks. We generate global snapshot features $\mathcal{F}_{\mathbf{gs}} \in \mathbb{R}^{t \times h \times w \times d}$ and $k$ local snapshot features $\mathcal{F}_{\mathbf{{ls}_{k}}} \in \mathbb{R}^{t \times h \times w \times d}$. The local snapshots are utilized to enhance the prediction derived solely from the global snapshot. We perform global average pooling on the spatial dimensions $h$ and $w$ of the snapshot features to obtain $\mathbb{R}^{t \times {h_{p}} \times {w_{p}} \times d}$ features, where ${h_{p}}$, ${w_{p}}$, $d$ are $4$, $4$, and $2048$ respectively. We flatten these features to $\mathbb{R}^{t{h_{p}}{w_{p}} \times d}$.

\subsubsection{Global and Local Snapshot Processing}
Similar to TemPr~\cite{stergiou2023wisdom}, we employ a transformer~\cite{attentionisallyouneed} model, denoted as $\phi_{i}$ (where $i \in [\mathbf{ls}_{1}, \mathbf{ls}_{2} \dots \mathbf{ls}_{k}]$ or $i = \mathbf{gs}$), to independently process the snapshot features $\mathcal{F}_{i}$. $\phi_{i}$ consists of $l$ layers of self-attention and feed-forward neural networks, generating local snapshot features $\mathcal{F}^{'}_{\mathbf{ls}_{k}}$ (for $i \in [1, k]$) and global snapshot features $\mathcal{F}^{'}_{\mathbf{gs}}$. Afterwards, we apply a bottleneck layer to reduce the feature dimension from $2048$ to $128$.

\subsubsection{Snapshot-Centric Attention (SCA) Module}
The features extracted from each of the $k$ local spatio-temporal snapshots of dimension $\mathbb{R}^{th_{p}w_{p} \times d}$ provide valuable scene information. However, distinctive cues related to operative difficulty in one snapshot might be unavailable to others, hindering the final prediction as snapshots lack mutual context. For instance, intraoperative findings such as \textit{adhesions covering more than 50\%} of the gallbladder might be visible in one camera view but not in another. To create a richer scene representation, these local snapshots need to interact. We propose a Snapshot-Centric Attention (SCA) module to address this issue. SCA uses inter-snapshot attention to enable semantic transfer between the $k$ spatio-temporal snapshots transforming features from $\mathcal{F}^{'}$ to $\mathcal{F}^{''} \in \mathbb{R}^{t{h_{p}}{w_{p}} \times d}$. This facilitates context-aware feature refinement, ensuring critical visual cues are shared for early LCOD assessment.

\begin{table*}[t!]
\centering
\caption{CholeScore phase list with mean $\pm$ std of the duration in seconds.}
\label{table:phase_distribution}
\begin{tabular}{c@{\hspace{0.6cm}}c@{\hspace{0.6cm}}c@{\hspace{0.6cm}}c@{\hspace{0.6cm}}c} 
\toprule
\textbf{ID} & \textbf{Phase} & \textbf{Duration (s)}\\
\midrule
P1 & Trocar placement and preparation & $798 \pm 1054$\\
P2 & Hepatocystic triangle (HCT) dissection & $1117 \pm 980$ \\
P3 & Clipping and cutting & $251 \pm 608$\\
P4 & Gallbladder bed dissection & $616 \pm 450$\\
P5 & Gallbladder packaging, extraction, cleaning and coagulation & $758 \pm 555$\\
P6 & Subtotal cholecystectomy & $1008 \pm 412$\\
\bottomrule
\end{tabular}
\end{table*}

\subsubsection{Input Pipeline and Loss Objective}
To enable batch processing of snapshots with variable $w$, we encode snapshot features $\mathbb{R}^{t{h_p}{w_p} \times d}$ with $w$ using a single-layer MLP, creating time-conditioned features. We apply a shared single-layer MLP to all snapshots, transforming the features from $\mathbb{R}^{t{h_p}{w_p} \times d}$ to $\mathbb{R}^C$. The per-snapshot predictions, $\hat{y}_{i} \in \mathbb{R}^C$, where $i \in [\mathbf{ls}_{1}, \mathbf{ls}_{2} \dots \mathbf{ls}_{k}]$ or $i = \mathbf{gs}$) are averaged to output a single class probability vector $\hat{y}$ for each window $w$. SurgPrOD is trained using cross-entropy loss as shown in Equation~\ref{loss_eq}:
\begin{equation}
\label{loss_eq}
L = - \sum_{b=1}^{B} \sum_{c=1}^{C} y_{b,c} \log(\hat{y}_{b,c}),
\end{equation}
where $B$ is batch size, $C$ is class count, $y_{b,c}$ is the binary ground truth, and $\hat{y}_{b,c}$ is the predicted probability for sample $b$ in class $c$.

\section{Experiments}
\subsection{Dataset}
The dataset was collected within the “$5$-second rule”~\cite{mascagni2021intraoperative} clinical trial enrolling adult patients undergoing elective LC for benign conditions at Nouvel Hopital Civil (Strasbourg, France) between November 2017 and November 2019. To make the dataset more treatable yet representative, the $343$ consecutive cases collected in the study were ranked by video duration and stratified random sampling was applied to select $25$ cases per quartile, resulting in a dataset of $100$ LC videos recorded at $25$ fps. Next, videos were temporally segmented according to surgical phases as in Cholec80~\cite{twinanda2016endonet}. Table~\ref{table:phase_distribution} lists observed phases with the mean duration (in seconds) and standard deviation. Subtotal cholecsystectomy is a bailout procedure and replaces \textit{clipping and cutting} phase in $4$ LC videos. Three independent clinicians with varying levels of surgical expertise annotated each phase with intraoperative findings included in the most validated IOAS available in the surgical literature - Parkland grading scale (PGS)~\cite{madni2018parkland}, Sugrue (S)~\cite{sugrue2015grading}, and Nassar (N)~\cite{nassar1995laparoscopic}. Each intraoperative finding was annotated as present, absent, or not assessable on MOSaiC~\cite{mazellier2023mosaic} annotation platform. This results in a sparse video-level annotation, lacking precise temporal information about the occurrence.
In this work, we focus only on the overall video-level LCOD assessment. The inter-rater agreement (Cohen's kappa) of the annotations was $72\%$ for PGS, $67\%$ for N, and $66\%$ for S.
We extract the frames at 1fps resulting in a total of $350$k frames. For each video, we increase the observation window $w$ (in minutes) from $1$ to $w_{max}$, where $w_{max}$ is set to $18$ (shortest video duration).

\subsection{Splits and Evaluation Metrics}
\begin{figure}[!h]
\centering
\includegraphics[width=0.49\textwidth]{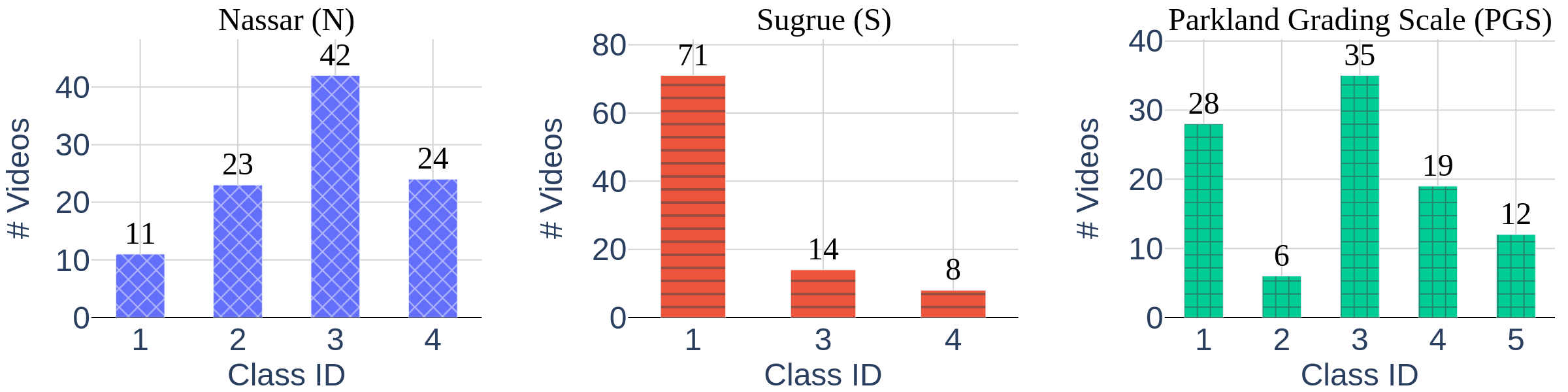}
\caption{Class Distribution: Parkland grading scale (P), Nassar (N), and Sugrue (S).}\label{fig:scale_class_dist}
\end{figure}

We perform majority voting across the three raters to obtain a single LCOD score per video. Next, we generate train-validation-test splits for each IOAS independently. We refer to grades in intraoperative assessment scales (typically 1-5) as classes, aligning with deep learning terminology. PGS, S, N contains $5$, $6$, $4$ LCOD classes, respectively. However, in Sugrue (S), the number of videos with class ID $2$, $5$ and $6$ is insufficient for creating splits; thus we only use videos with class ID $1$, $3$, and $4$. Finally, we apply stratified sampling to generate splits, as illustrated in Figure~\ref{fig:scale_class_dist}.
The Parkland grading scale (PGS)~\cite{madni2018parkland}, with five classes, is divided into $52$ training, $16$ validation, and $32$ test videos. For the Sugrue (S)~\cite{sugrue2015grading} scale, with three classes, we use $48$ training, $15$ validation, and $30$ test videos. The Nassar (N)~\cite{nassar1995laparoscopic} scale, with four classes, is split into $53$ training, $17$ validation, and $30$ test videos.

We treat early LCOD assessment as a multi-class classification task. 
We report top-1 accuracy, F1-score, and Quadratic-weighted Cohen's Kappa (QWK), first averaged over all observation windows $w$ for each video, and then averaged across all videos.
However, we observe that these metrics do not fully capture the model's capability to provide accurate and stable predictions early in the procedure, as they treat all predictions equally, regardless of $w$. To address this limitation, we introduce the Earliness-Stability (ES) metric described in Equation~\ref{es_metric}, which is illustrated by a comparison with traditional metrics in Figure~\ref{fig:surgprod} (right). The ES metric, denoted $ES(n,v)$, considers two key aspects of the model performance for a given video $v$ and window step size $n$: ($1$) Earliness: how early the model predicts the correct class with confidence exceeding $\tau$ (Hit($\cdot$)), ($2$) Stability: whether the correct prediction persists from $w+1$ to $w + n - 1$ (assessed by $S(w,n)$).

\begin{figure*}[h]
\begin{equation}
\begin{gathered}
\begin{aligned}
ESV(n) &= \frac{1}{|\mathcal{V}|} \sum_{v \in \mathcal{V}} ES(n,v), \text{ where }
ES(n,v) = \begin{cases}
\frac{w_{\text{max}} - w + S(w,n)}{w_{\text{max}}} & \text{if } \exists w: \text{Hit}(w) \\
0 & \text{otherwise}\end{cases}, \\
S(w,n) &= \frac{1}{n} \sum_{j=w+1}^{\min(w+n-1, P)} \mathds{1}\{\text{Hit}(j)\},
\text{ and } \text{Hit}(k) = (c_k > \tau) \land (\hat{y}_k = y_k). \\
\end{aligned} \\
meanES = \frac{1}{3} (ESV(1) + ESV(3) + ESV(5)),
\end{gathered}
\label{es_metric}
\end{equation}
\end{figure*}

where $\mathcal{V}$ is the set of all videos, $\mathds{1}[\cdot]$ is the indicator function, $P$ is the number of windows per video, $c_k = \text{Softmax}(x_k)$ is the prediction confidence, $\hat{y}_k = \arg\max(c_k)$ is the predicted label, and $y_k$ is the true label at window $k$. Specifically, for a given video $v$ and observation windows $1$ to $w_{max}$, ES identifies the earliest window $w$ where Hit($w$) occurs. To ensure robust early predictions by mitigating the risk of relying on a single, potentially spurious, correct prediction, ES assesses prediction stability from $w+1$ to $w+n-1$. The per-video ES value, denoted as $ES(n, v)$, is then averaged across all videos $\mathcal{V}$ to obtain the ES over Videos metric, $ESV(n)$. Computing ESV for $n \in \left\{1, 3, 5 \right\}$ evaluates stability at increasing short-term time horizons, capturing different levels of persistence. These $ESV(n)$ values are then averaged to obtain the meanES metric, which is bounded between [$0, 1$).

\subsection{Implementation Details}
We resize frames to $224\times224$ and apply RandAugment~\cite{cubuk2020randaugment} for training augmentation. We sample $\textit{t}=8$ frames randomly during training and uniformly during evaluation. For the snapshot feature extractor $\phi$, we use a $4$-layer transformer with $4$ number of heads. We utilize $1$ block of SCA for PGS, S and $2$ blocks for N. We employ a 1-layer MLP to generate class logits. We train SurgPrOD end-to-end with AdamW optimizer using $1e^{-5}$ learning rate, $5e^{-2}$ weight decay for $30$ epochs. We use $8$ as batch size and decay the learning rate by $0.1$ at epoch $10$ and $20$. SurgPrOD is implemented in PyTorch (version $2.1.1$) using MMaction2~\cite{2020mmaction2} framework (version $1.2.0$). We train models on Nvidia A100 GPU (CUDA $12.1$) and tune model hyperparameters on validation videos for each IOAS independently on meanES metric with $\tau$ set default to $0.70$.

\subsection{Results}
We present our results in Table~\ref{table:main_results} for three IOAS: Parkland (PGS), Nassar (N), and Sugrue (S). We establish three baselines for fair comparison with SurgPrOD: a random baseline with predictions obtained from a multivariate normal distribution for each scale independently; a Vision Transformer (ViT-S)~\cite{vit} with VideoMAE pretrained weights; and an image-based EndoViT~\cite{batic2024endovit} with pretrained weights from the public HuggingFace repository. To provide an upper bound (Human Performance), we use majority-voted LCOD annotations as a proxy for ground truth. We treat each rater's individual annotation as a prediction, computing accuracy, F1, and QWK for each. As the meanES metric requires access to observation window during prediction, we do not compute it. The reported metrics are averaged across the three raters.
We consider three variants of SurgPrOD: (G) with only global snapshot, (GL) with global and \textit{k} local snapshots, and (GL-SCA) with snapshot-centric attention on \textit{k} local snapshots. We fine-tune the ViT-S and EndoViT backbone weights on the early LCOD prediction task using the variants G, GL and GL-SCA. We report the best results achieved with the GL-SCA setting. We observe the best performance with \textit{k} set to $2$. Across all three IOAS, SurgPrOD consistently outperforms baselines methods. SurgPrOD with local snapshots and SCA improves over baselines in Top1-Acc by $5.57$, $24.33$, and $5.96$ percentage points (pp) in PGS, S, and N respectively. Similar trends in F1-score and QWK metrics further demonstrate SurgPrOD's effectiveness in LCOD prediction. For the meanES metric, SurgPrOD with SCA achieves gains over baselines of $0.32$ points in PGS, $0.23$ points in S, and $0.22$ points in N. This demonstrates that SurgPrOD with SCA is not only better at making correct early predictions but also more stable compared to the baselines. Multiple local snapshots capture fine-grained temporal changes, while SCA enables effective information exchange between snapshots. This emphasizes relevant features and suppress irrelevant cues for LCOD prediction. Models marked with
† (poor performance at $\tau=0.7$) were evaluated at $\tau=0.5$. The GL-SCA variant, which utilizes \textit{k} local snapshots and a global snapshot (totaling $24$ frames when $k=2$), leads to a higher memory footprint for these models.

\begin{table}[h]
\begin{center} 
\caption{Results on early LCOD assessment. Models marked with $\dagger$ are evaluated on meanES with $\tau \ge 0.5$. Mem: peak memory footprint (Nvidia A100 GPU, batch size $8$).}
\label{table:main_results}
\begin{threeparttable}
\resizebox{\columnwidth}{!}{%
\begin{tabular}{@{\extracolsep{\fill}}c c cc cc c}
\toprule
Scale & Methods & top1-Acc & F1 & QWK & meanES & Mem \\

\midrule
\multirow{6}{*}{PGS}
& Human Performance & $83.70$ & $81.15$ & $0.930$ & - & -\\
& Random Baseline & $17.12$ & $16.42$ & $0.018$ & $0.14$ & -\\
& ViT-S (GL-SCA) & $30.87$ & $26.59$ & $0.422$ & $0.29$ & $10.03$GB \\
& EndoViT~\cite{batic2024endovit} (GL-SCA) $\dagger$ & $23.73$ & $21.47$ & $0.252$ & $0.14$ & $22.95$GB\\
& SurgPrOD (G) & $28.88$ & $25.76$ & $0.418$ & $0.58$ & $6.90$GB \\
& SurgPrOD (GL) & $32.33$ & $29.06$ & $0.500$ & $0.57$ & $19.79$GB \\
& SurgPrOD (GL-SCA) & $\textbf{36.44}$ & $\textbf{35.87}$ & $\textbf{0.590}$  & $\textbf{0.61}$ & $19.79$GB\\

\midrule

\multirow{6}{*}{S}
& Human Performance & $90.03$ & $87.60$ & $0.802$ & - & - \\
& Random Baseline & $39.73$ & $32.09$ & $0.055$ & $0.51$ & - \\
& ViT-S (GL-SCA) & $30.68$ & $29.69$ & $0.286$ & $0.49$ & $10.03$GB \\
& EndoViT~\cite{batic2024endovit} (GL-SCA) $\dagger$ & $34.13$ & $32.50$ & $0.025$ & $0.64$ & $22.95$GB \\
& SurgPrOD (G) & $49.28$ & $48.38$ & $0.335$ & $0.81$ & $6.90$GB \\
& SurgPrOD (GL) & $51.98$ & $53.12$ & $0.331$ & $0.84$ & $19.79$GB \\
& SurgPrOD (GL-SCA) & $\textbf{64.06}$ & $\textbf{64.88}$ & $\textbf{0.512}$ & $\textbf{0.87}$ & $19.79$GB\\
\midrule

\multirow{6}{*}{N}
& Human Performance & $70.90$ & $72.21$ & $0.794$ & - & - \\
& Random Baseline & $26.15$ & $23.40$ & $0.042$ & $0.25$ & - \\
& ViT-S (GL-SCA) & $35.51$ & $31.61$ & $0.249$ & $0.30$ & $10.03$GB \\
& EndoViT~\cite{batic2024endovit} (GL-SCA) $\dagger$ & $19.20$ & $17.20$ & $-0.118$ & $0.20$ & $22.95$GB \\
& SurgPrOD (G) & $36.00$ & $36.56$ & $\textbf{0.406}$ & $0.49$ & $6.90$GB \\
& SurgPrOD (GL) & $38.53$ & $31.06$ & $0.396$ & $0.48$ & $19.79$GB \\
& SurgPrOD (GL-SCA) & $\textbf{41.47}$ & $\textbf{42.37}$ & $0.307$ & $\textbf{0.52}$ & $19.79$GB \\

\bottomrule
\end{tabular}
} 
\end{threeparttable}
\end{center}
\end{table}

\subsection{Ablations}
\subsubsection{Impact of frame count in snapshots:} 
We analyze the impact of increasing the number of frames in snapshots. Figure~\ref{fig:ablation_results}(a) shows that performance improves up to $t = 8$ frames, after which it diminishes. This implies that increasing the number of frames in local snapshots, which are more localized than the global snapshot, often captures similar visual cues and risks leading to overcompensation by the model.

\subsubsection{Impact of number of local snapshots \textit{k}:} 
Figure~\ref{fig:ablation_results}(b) shows that increasing \textit{k} from $1$ to $2$ improves meanES, suggesting multiple snapshots effectively capture temporal dynamics. However, beyond $2$ snapshots, performance diminishes, as this introduces more redundant information.

\subsubsection{SCA vs No SCA:} 
\begin{figure}[h]
\centering
\includegraphics[width=0.48\textwidth]{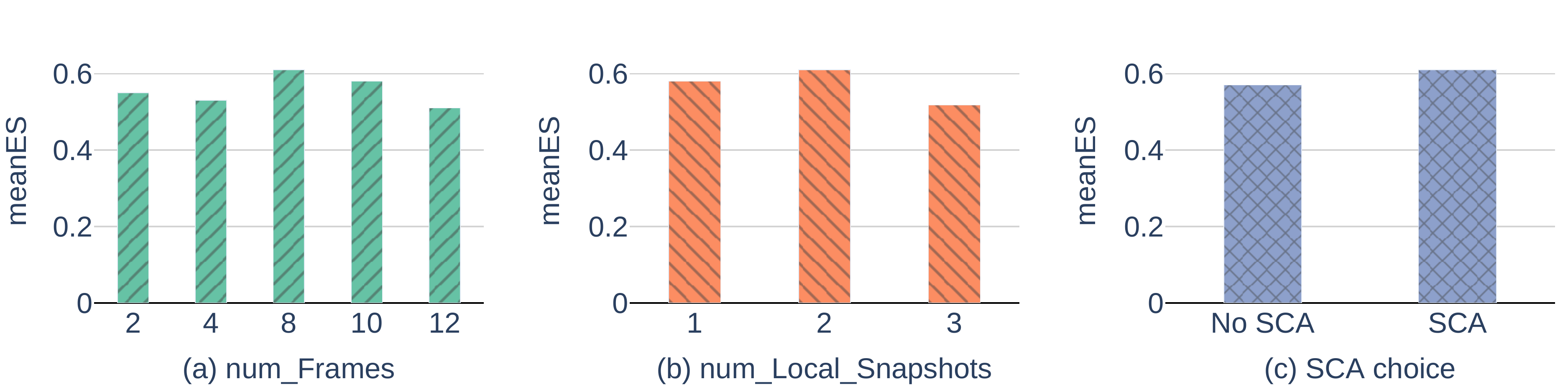} 
\caption{Ablation studies on SurgPrOD.}\label{fig:ablation_results}
\end{figure}

Figure~\ref{fig:ablation_results}(c) shows that SurgPrOD with snapshot-centric attention (SCA) module improves inter-snapshot contextual features, enabling enhancing early prediction. Without SCA, the model lacks inter-snapshot awareness necessary for identifying operative difficulty cues.

\subsection{Qualitative Analysis}
\subsubsection{Visualization of LCOD prediction across \textit{w}:} 
\begin{figure}[h]
\centering
\includegraphics[width=0.47\textwidth]{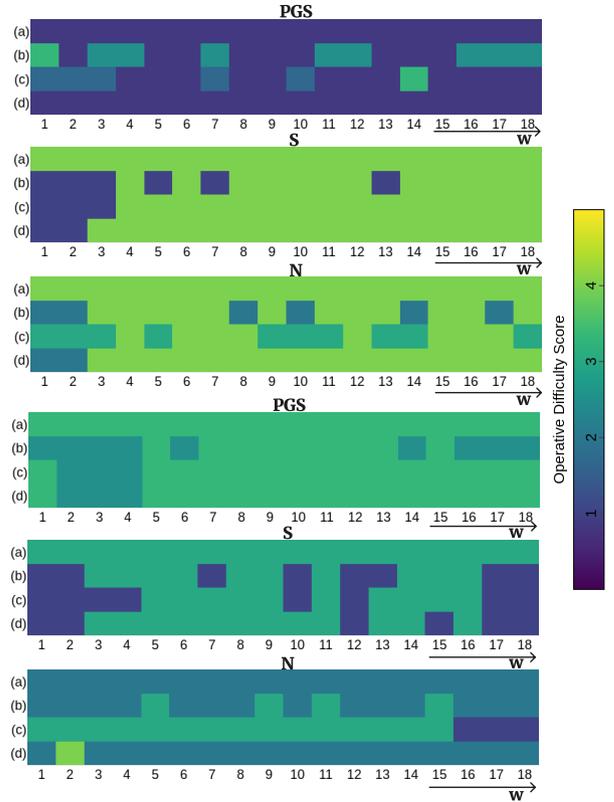}
\caption{Visualization of model predictions on randomly sampled $6$ test videos. (a) Ground truth LCOD label. SurgPrOD with (b) Global (G) only snapshot, (c) Global and local snapshot (GL), and (d) Snapshot-centric attention module (GL + SCA).} \label{fig:eod_overall_vis}
\end{figure}

Figure~\ref{fig:eod_overall_vis} shows that for all three IOAS, SurgPrOD with the SCA module surpass its counterparts. This shows that SCA reduces fragmentation, observed with independent snapshot feature processing. In some cases, SurgPrOD mispredicts for higher \textit{w}, likely due to noise from sampling temporally distant features across the large observation window.

\subsubsection{Where does SCA focus on?} 
\begin{figure}[h]
\centering
\includegraphics[width=0.45\textwidth]{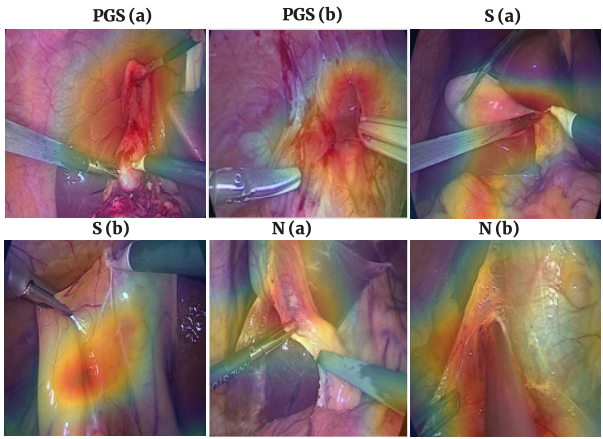}
\caption{Visualization of SCA attention map for Parkland grading scale (PGS), Nassar (N), and Sugrue (S). Models tend to focus on both tools and anatomical structures.}\label{fig:sca_heatmap}
\end{figure}

We visualize the attention map (Figure~\ref{fig:sca_heatmap}) to highlight regions of maximum confidence. SCA primarily focuses on tool-tissue interaction regions, as shown in PGS(a) and PGS(b). In some cases, it also emphasizes relevant anatomical structures. For instance, in S(b), SCA partially focuses on gallbladder adhesions and visceral fat, both key indicators of operative difficulty in Sugrue.

\section{Conclusion}
In this work, we introduce the novel task of early operative difficulty assessment in laparoscopic cholecystectomy. We exploit robust video clip features combined with our novel local and global snapshot-based feature selection and a snapshot-centric attention module acting on local snapshots to predict operative difficulty. To conduct our experiments, we introduce the CholeScore dataset, featuring sparse annotations from three intraoperative LCOD assessment scales. Our results demonstrate that both local and global snapshots are necessary for accurate early prediction. To enable fair comparison, we introduce an earliness-stability metric conditioned on time. Future work will explore integrating preoperative features to further enhance LCOD prediction.


\section{Acknowledgements}
This work was supported by French state funds managed by the ANR within the National AI Chair program under Grant ANR-20-CHIA-0029-01 (Chair AI4ORSafety) and within the Investments for the future program under Grant ANR-10-IAHU-02 (IHU Strasbourg). It was granted access to the GENCI-IDRIS (Grant AD011013710R2).

\noindent{\bf Ethical approval}
The University of Strasbourg's Ethics Committee approved data collection for the “SafeChole – Surgical Data Science for Safe Laparoscopic Cholecystectomy” study (ID F20200730144229). In compliance with the French MR004 reference framework, all data was obtained with patient consent and anonymized by removing identifying information.



\noindent{\bf Competing interests} The authors declare no conflict of interest.

\noindent{\bf Informed consent} This manuscript does not contain any patient data.

\noindent{\bf Code availability} Source code will be provided at \url{https://github.com/CAMMA-public/cholescore}.

\bibliographystyle{sn-basic}
\bibliography{sn-bibliography}

\begin{thebibliography}{23}
\providecommand{\natexlab}[1]{#1}
\providecommand{\url}[1]{{#1}}
\providecommand{\urlprefix}{URL }
\providecommand{\doi}[1]{\url{https://doi.org/#1}}
\providecommand{\eprint}[2][]{\url{#2}}

\bibitem[{Bati{\'c} et~al(2024)Bati{\'c}, Holm, {\"O}zsoy, Czempiel, and Navab}]{batic2024endovit}
Bati{\'c} D, Holm F, {\"O}zsoy E, et~al (2024) Endovit: pretraining vision transformers on a large collection of endoscopic images. IJCARS 19(6):1085--1091

\bibitem[{Contributors(2020)}]{2020mmaction2}
Contributors M (2020) Openmmlab's next generation video understanding toolbox and benchmark. \url{https://github.com/open-mmlab/mmaction2}

\bibitem[{Cubuk et~al(2020)Cubuk, Zoph, Shlens, and Le}]{cubuk2020randaugment}
Cubuk ED, Zoph B, Shlens J, et~al (2020) Randaugment: Practical automated data augmentation with a reduced search space. In: CVPR workshops, pp 702--703

\bibitem[{Dosovitskiy et~al(2021)Dosovitskiy, Beyer et~al}]{vit}
Dosovitskiy A, Beyer L, et~al (2021) An image is worth 16x16 words: Transformers for image recognition at scale. In: ICLR 2021

\bibitem[{Kannan et~al(2019)Kannan, Yengera, Mutter, Marescaux, and Padoy}]{kannan2019future}
Kannan S, Yengera G, Mutter D, et~al (2019) Future-state predicting lstm for early surgery type recognition. IEEE TMI 39(3):556--566

\bibitem[{Lavanchy et~al(2024)Lavanchy, Ramesh, Dall’Alba, Gonzalez, Fiorini, M\"{u}ller-Stich, Nett, Marescaux, Mutter, and Padoy}]{Lavanchy2024}
Lavanchy JL, Ramesh S, Dall’Alba D, et~al (2024) Challenges in multi-centric generalization: phase and step recognition in roux-en-y gastric bypass surgery. IJCARS

\bibitem[{Madni et~al(2018)Madni, Leshikar, Minshall et~al}]{madni2018parkland}
Madni TD, Leshikar DE, Minshall, et~al (2018) The parkland grading scale for cholecystitis. The American Journal of Surgery 215(4):625--630

\bibitem[{Maier-Hein et~al(2022)Maier-Hein, Eisenmann, Sarikaya, M{\"a}rz, Collins, Malpani, Fallert, Feussner, Giannarou, Mascagni et~al}]{maier2022surgical}
Maier-Hein L, Eisenmann M, Sarikaya D, et~al (2022) Surgical data science--from concepts toward clinical translation. MedIA 76:102306

\bibitem[{Mascagni et~al(2021)Mascagni, Rodr{\'\i}guez-Luna, Urade, Felli, Pessaux, Mutter, Marescaux, Costamagna, Dallemagne, and Padoy}]{mascagni2021intraoperative}
Mascagni P, Rodr{\'\i}guez-Luna MR, Urade T, et~al (2021) Intraoperative time-out to promote the implementation of the critical view of safety in laparoscopic cholecystectomy: a video-based assessment of 343 procedures. Journal of the American College of Surgeons 233(4):497--505

\bibitem[{Mascagni et~al(2022)Mascagni, Vardazaryan, Alapatt, Urade, Emre, Fiorillo, Pessaux, Mutter, Marescaux et~al}]{mascagni2022artificial}
Mascagni P, Vardazaryan A, Alapatt D, et~al (2022) Artificial intelligence for surgical safety: automatic assessment of the critical view of safety in laparoscopic cholecystectomy using deep learning. Annals of surgery 275(5):955--961

\bibitem[{Mazellier et~al(2023)Mazellier, Boujon, Bour-Lang, Erharhd, Waechter, Wernert, Mascagni, and Padoy}]{mazellier2023mosaic}
Mazellier JP, Boujon A, Bour-Lang M, et~al (2023) Mosaic: a web-based platform for collaborative medical video assessment and annotation. arXiv preprint arXiv:231208593

\bibitem[{Murali et~al(2023)Murali, Alapatt, Mascagni, Vardazaryan, Garcia, Okamoto, Mutter, and Padoy}]{murali2023latent}
Murali A, Alapatt D, Mascagni P, et~al (2023) Latent graph representations for critical view of safety assessment. IEEE TMI pp 1--1

\bibitem[{Nassar et~al(1995)Nassar, Ashkar, Mohamed, and Hafiz}]{nassar1995laparoscopic}
Nassar A, Ashkar K, Mohamed A, et~al (1995) Is laparoscopic cholecystectomy possible without video technology? Minimally Invasive Therapy 4(2):63--65

\bibitem[{Nwoye et~al(2022)Nwoye, Yu, Gonzalez, Seeliger, Mascagni, Mutter, Marescaux, and Padoy}]{nwoye2021rendezvous}
Nwoye CI, Yu T, Gonzalez C, et~al (2022) Rendezvous: Attention mechanisms for the recognition of surgical action triplets in endoscopic videos. MedIA 78:102433

\bibitem[{Ramesh et~al(2023)Ramesh, Srivastav, Alapatt, Yu, Murali, Sestini, Nwoye, Hamoud, Sharma, Fleurentin et~al}]{ramesh2023dissecting}
Ramesh S, Srivastav V, Alapatt D, et~al (2023) Dissecting self-supervised learning methods for surgical computer vision. MedIA p 102844

\bibitem[{Sharma et~al(2023{\natexlab{a}})Sharma, Nwoye, Mutter, and Padoy}]{sharma2022rendezvous}
Sharma S, Nwoye CI, Mutter D, et~al (2023{\natexlab{a}}) Rendezvous in time: an attention-based temporal fusion approach for surgical triplet recognition. IJCARS 18(6):1053--1059

\bibitem[{Sharma et~al(2023{\natexlab{b}})Sharma, Nwoye, Mutter, and Padoy}]{sharma2023surgical}
Sharma S, Nwoye CI, Mutter D, et~al (2023{\natexlab{b}}) Surgical action triplet detection by mixed supervised learning of instrument-tissue interactions. MICCAI

\bibitem[{Stergiou and Damen(2023)}]{stergiou2023wisdom}
Stergiou A, Damen D (2023) The wisdom of crowds: Temporal progressive attention for early action prediction. In: CVPR, pp 14709--14719

\bibitem[{Sugrue et~al(2015)Sugrue, Sahebally, Ansaloni, and Zielinski}]{sugrue2015grading}
Sugrue M, Sahebally SM, Ansaloni L, et~al (2015) Grading operative findings at laparoscopic cholecystectomy-a new scoring system. World Journal of Emergency Surgery 10:1--8

\bibitem[{Twinanda et~al(2016)Twinanda, Shehata, Mutter, Marescaux, De~Mathelin, and Padoy}]{twinanda2016endonet}
Twinanda AP, Shehata S, Mutter D, et~al (2016) Endonet: a deep architecture for recognition tasks on laparoscopic videos. IEEE TMI 36(1):86--97

\bibitem[{Vannucci et~al(2022)Vannucci, Laracca, Mercantini, Perretta, Padoy, Dallemagne, and Mascagni}]{vannucci2022statistical}
Vannucci M, Laracca GG, Mercantini P, et~al (2022) Statistical models to preoperatively predict operative difficulty in laparoscopic cholecystectomy: a systematic review. Surgery 171(5):1158--1167

\bibitem[{Vaswani et~al(2017)Vaswani, Shazeer, Parmar, Uszkoreit, Jones, Gomez, Kaiser, and Polosukhin}]{attentionisallyouneed}
Vaswani A, Shazeer N, Parmar N, et~al (2017) Attention is all you need. NeurIPS 30

\bibitem[{Vercauteren et~al(2019)Vercauteren, Unberath, Padoy, and Navab}]{vercauteren2019cai4cai}
Vercauteren T, Unberath M, Padoy N, et~al (2019) Cai4cai: the rise of contextual artificial intelligence in computer-assisted interventions. Proceedings of the IEEE 108(1):198--214

\end{thebibliography}

\end{document}